\documentclass{article}
\usepackage{ijcai24}

\usepackage{times}
\usepackage{soul}
\usepackage{url}
\usepackage[hidelinks]{hyperref}
\usepackage[utf8]{inputenc}
\usepackage[small]{caption}
\usepackage{graphicx}
\usepackage{amsmath}
\usepackage{amsthm}
\usepackage{booktabs}
\usepackage{algorithm}
\usepackage{algorithmic}
\usepackage[switch]{lineno}
\usepackage{xcolor}
\usepackage{kotex}
\usepackage{microtype}
\usepackage{IEEEtrantools}

\usepackage{siunitx}
\usepackage{tabularx}
\usepackage{varwidth}
\usepackage{multirow}
\usepackage{makecell}
\usepackage{caption}
\usepackage{subcaption}

\theoremstyle{definition} \newtheorem{definition}{Definition}

\title{Contextual Sprint Classification in Soccer Based on Deep Learning}

\author{
Hyunsung Kim$^1$\and
Gun-Hee Joe$^2$\and
Jinsung Yoon$^1$\and
Sang-Ki Ko$^{1,3}$\\
\affiliations
$^1$Fitogether Inc., Republic of Korea\\
$^2$Kangwon National University, Republic of Korea\\
$^3$University of Seoul, Republic of Korea\\
\emails
\{hyunsung.kim, jinsung.yoon, sangki.ko\}@fitogether.com,\\
gun926777@kangwon.ac.kr, sangkiko@uos.ac.kr
}

\begin{document}

\maketitle

\begin{abstract}
The analysis of high-intensity runs (or sprints) in soccer has long been a topic of interest for sports science researchers and practitioners. In particular, recent studies suggested contextualizing sprints based on their tactical purposes to better understand the physical-tactical requirements of modern match-play. However, they have a limitation in scalability, as human experts have to manually classify hundreds of sprints for every match. To address this challenge, this paper proposes a deep learning framework for automatically classifying sprints in soccer into contextual categories. The proposed model covers the permutation-invariant and sequential nature of multi-agent trajectories in soccer by deploying Set Transformers and a bidirectional GRU. We train the model with category labels made through the collaboration of human annotators and a rule-based classifier. Experimental results show that our model classifies sprints in the test dataset into 15 categories with the accuracy of \SI{77.65}{\%}, implying the potential of the proposed framework for facilitating the integrated analysis of soccer sprints at scale.
\end{abstract}

\section{Introduction}
As a vast amount of player tracking data begins to flood into the field of soccer, there is increasing attention to data-based analysis to provide relevant insights to domain participants \cite{ReinM16}. One of the applications that has quickly spread to practitioners is monitoring and managing players’ fitness based on physical metrics \cite{BuchheitS17}. Especially, the amount of high-intensity activity is considered to have an important impact on match outcome \cite{BradleyA18}, whose average running distance in a top-tier match has been significantly increasing as years go by \cite{BarnesAHBB14,BushBAHB15}. Thus, there is a growing interest in scrutinizing high-intensity runs (or sprints) in various aspects and making players robust enough to meet the physical demands of contemporary match-play \cite{BradleyA18}.

In addition, recent studies \cite{BradleyA18,JuDHELB22,JuLELB22,CaldbeckD23,JuDHELB23} went into focusing on ``why’’ the sprints occurred during a match rather than only on the total amount of physical demands. Consequently, they suggested subgrouping sprints according to their tactical purposes and analyzed the distribution of sprint categories for different roles of players. This context-aware analysis allowed practitioners to design more tailored training drills for each role or even player.

A limitation of the above approaches is that it is less scalable in practice because human experts have to manually annotate the tactical roles of players and the contextual categories of sprints for every match. Particularly, while several papers \cite{BialkowskiLCYSM14,ShawG19,KimKCYK22,BauerAS23} have proposed data mining techniques to detect players’ roles and their changes during a match, there have been very few systematic approaches to automatically classify sprints according to tactical intentions. Llana et al. \shortcite{LlanaBMF22} proposed a framework for categorizing sprints by the relative locations of their starting and ending points, but their naive classification did not satisfy the semantic granularity required in the aforementioned studies. (For example, two ``inside to wing'' sprints in their framework can have different tactical intentions, e.g., exploiting space, supporting a teammate, or pressing an opponent.) Considering that hundreds of sprints occur in every match, it is infeasible for practitioners to classify sprints on each occasion to identify appropriate physical-tactical demands.

To fill this gap, this paper proposes a deep learning framework for automatically classifying sprints in soccer into contextual categories. The model consists of Set Transformers \cite{LeeLKKCT19} for permutation-invariant representation of game contexts and a bidirectional GRU \cite{ChoVGBBSB14} for modeling their sequential nature. For data labeling, we adapt the taxonomy proposed by Ju et al. \shortcite{JuDHELB23} to have more quantitative classification criteria. With this taxonomy, we generate labels by human annotation with additional correction by a rule-based classifier and domain experts. Experimental results show that our model classifies sprints in the test dataset into 15 categories with the accuracy of \SI{77.65}{\%}. Lastly, we suggest some use cases of the categorized sprints in practical scenarios including estimation of physical-tactical match demands and similar play retrieval. The contribution of the proposed framework lies in connecting concepts in sports science with a data-driven approach, thereby lowering the cost of their extensive application to the field.

\section{Methodology}
This section elaborates on the details of two types of sprint classifiers. Specifically, Section \ref{se:detection} explains the process of detecting sprints from a player's speed signal. Then, Sections \ref{se:rule_engine} and \ref{se:deep_model} introduce a rule-based engine and a deep learning model, respectively, that classify the detected sprints using the trajectories of interacting players and the ball.

\subsection{Detecting Sprints from Soccer Match Data}
\label{se:detection}
Previous studies in the sports science field \cite{BradleyA18,JuLELB22,JuDHELB23,CaldbeckD23} defined a sprint as a movement where a player maintains a speed higher than a threshold value for a certain period. However, since force is proportional to acceleration, not speed, a sprint interval defined as above is usually discordant with the interval in which the player actually intends to sprint. Hence, one detected sprint interval may include multiple tactical actions \cite{JuDHELB23}, which hampers precise matching of sprints and their true intentions.

To more accurately and consistently find the ``intended'' sprint interval, we define a sprint based on acceleration similar to Llana et al. \shortcite{LlanaBMF22}. That is, we first define a \emph{run effort} as a series of movements in which a player starts to accelerate, reaches a peak speed, and decelerates. Then, we detect run efforts with peak speed higher than \SI{21}{\kilo\meter\per\hour} as \emph{sprints}.

\begin{figure}[htb]
    \centering
    \includegraphics[width=0.48\textwidth]{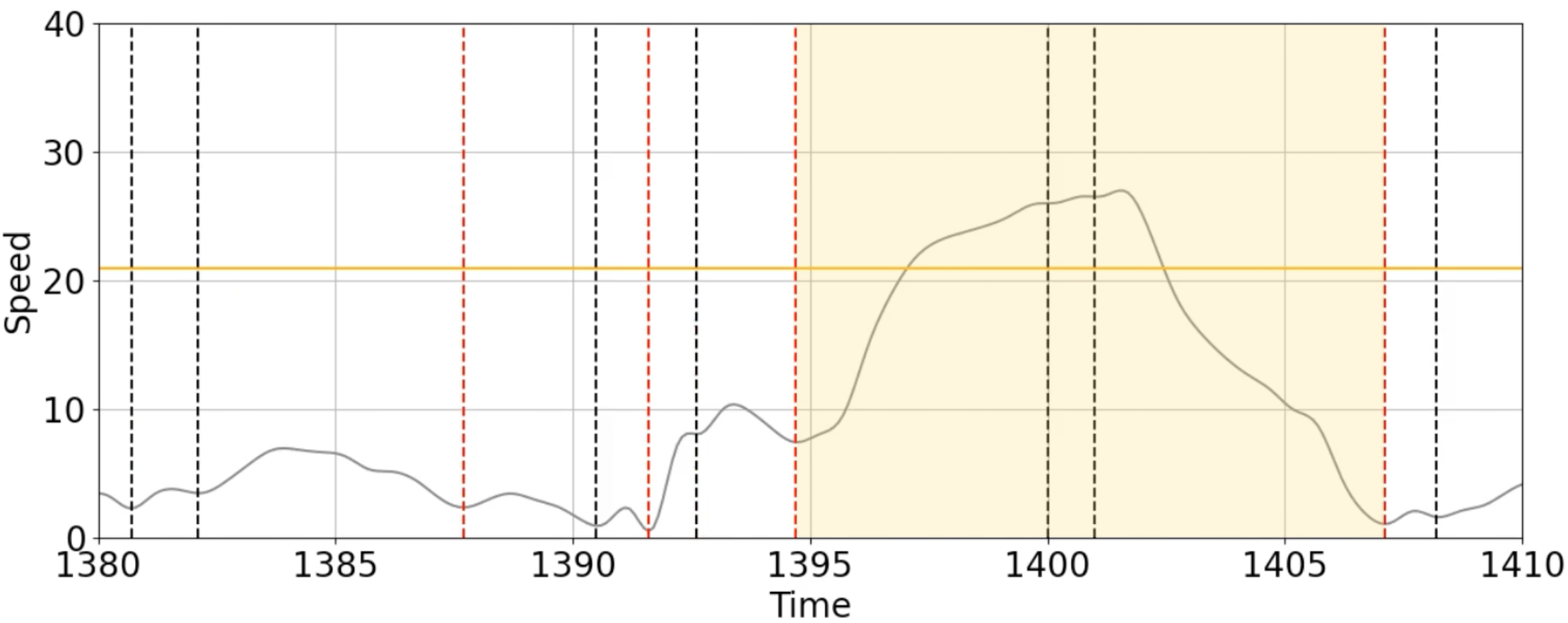}
    \caption{Speed plot of a player in a match. The dotted vertical lines indicate the borderlines between every deceleration and the following acceleration, among which selected cut-off points of run efforts are colored in red. The shaded area is a sprint whose peak speed exceeds \SI{21}{\kilo\meter\per\hour} (i.e., the yellow horizontal line).}
    \label{fig:speed_plot}
\end{figure}

It should be noted that if we detect every acceleration-deceleration pair as a run effort, a single effort may be incorrectly detected as multiple ones due to data fluctuation or the player's detailed movements. For example, the shaded area in Figure \ref{fig:speed_plot} consists of three acceleration-deceleration pairs, but it is indeed a single run effort of the player.

Thus, we set a threshold value $\tau$ for speed change to make this detection process robust to noise signals. More specifically, we only detect a valley as a valid cut-off point of run efforts if it follows a speed decrease by more than $\tau$ from the previous peak speed or precedes a speed increase by more than $\tau$ until the next peak. $\tau$ is empirically set to \SI{4}{\kilo\meter\per\hour}.

This acceleration-based detection makes each effort interval equal to a period the player actually makes an effort. In addition, this definition is more consistent in that even if the threshold value for peak speed changes, the length of each interval does not. Note that approaches~\cite{BradleyA18,JuDHELB22,JuLELB22,CaldbeckD23,JuDHELB23} that defined a sprint as an interval with speed higher than a threshold value are inherently sensitive to the value since the endpoints of a sprint are dependent to it.

\subsection{Quantifying the Taxonomy and a Rule-Based Classifying Engine}
\label{se:rule_engine}
\begin{figure}[bht]
    \centering
    \begin{subfigure}[t]{0.23\textwidth}
        \includegraphics[width=\textwidth]{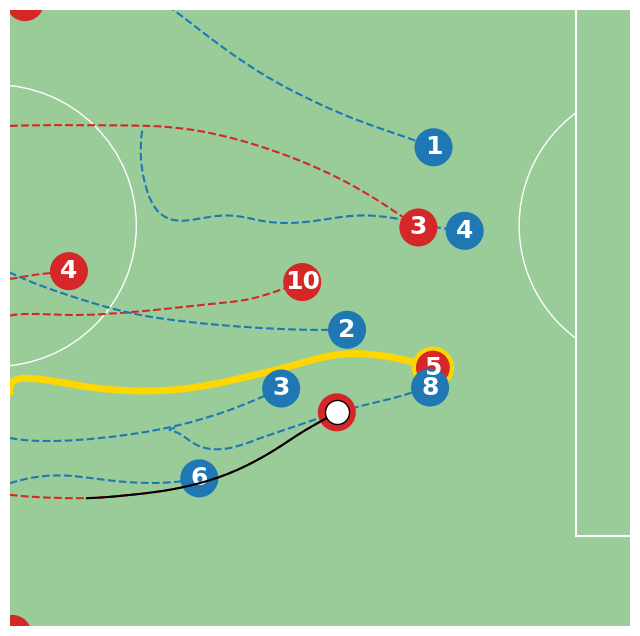}
        \caption{Penetration (PEN).}
        \label{fig:example_pen}
    \end{subfigure}
    \begin{subfigure}[t]{0.23\textwidth}
        \centering
        \includegraphics[width=\textwidth]{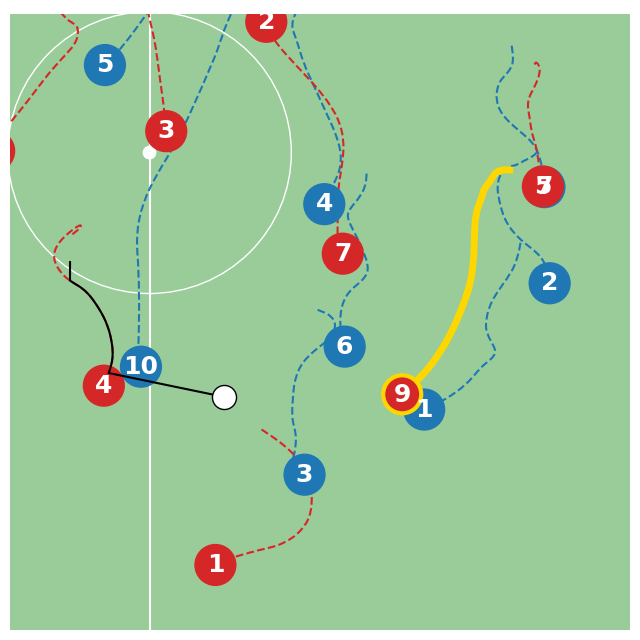}
        \caption{Move to Receive (MTR).}
        \label{fig:example_mtr}
    \end{subfigure}
    \begin{subfigure}[t]{0.23\textwidth}
        \includegraphics[width=\textwidth]{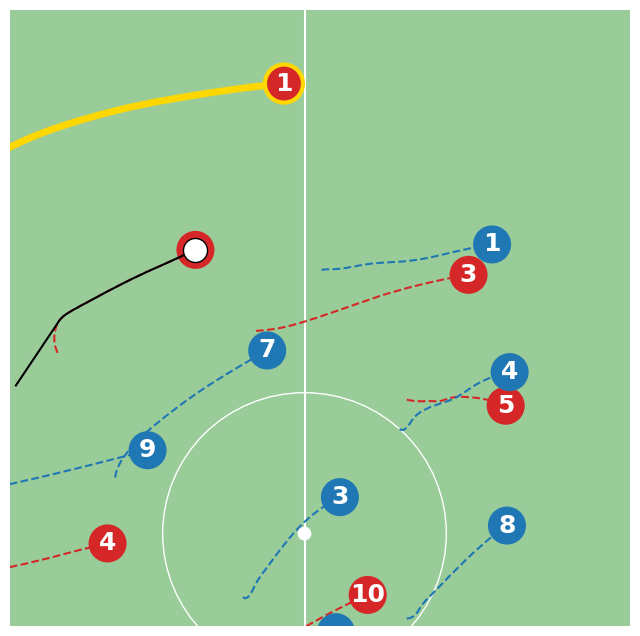}
        \caption{Overlapping (OVL).}
        \label{fig:example_ovl}
    \end{subfigure}
    \begin{subfigure}[t]{0.23\textwidth}
        \centering
        \includegraphics[width=\textwidth]{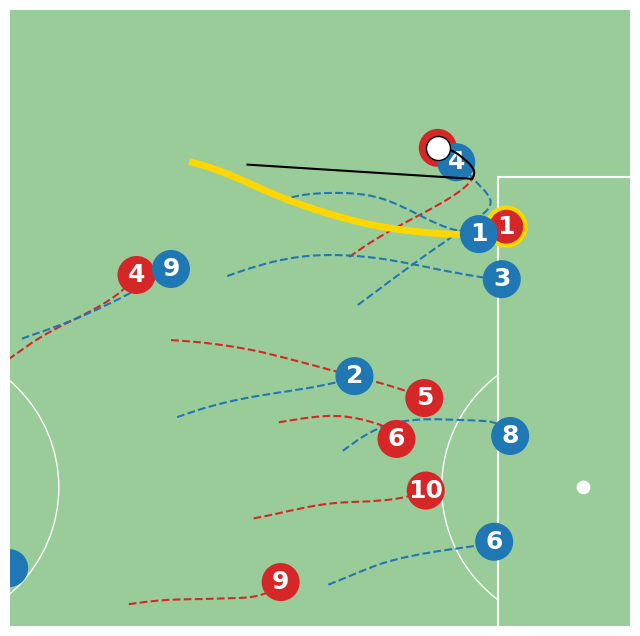}
        \caption{Underlapping (UNL).}
        \label{fig:example_unl}
    \end{subfigure}
    \begin{subfigure}[t]{0.23\textwidth}
        \centering
        \includegraphics[width=\textwidth]{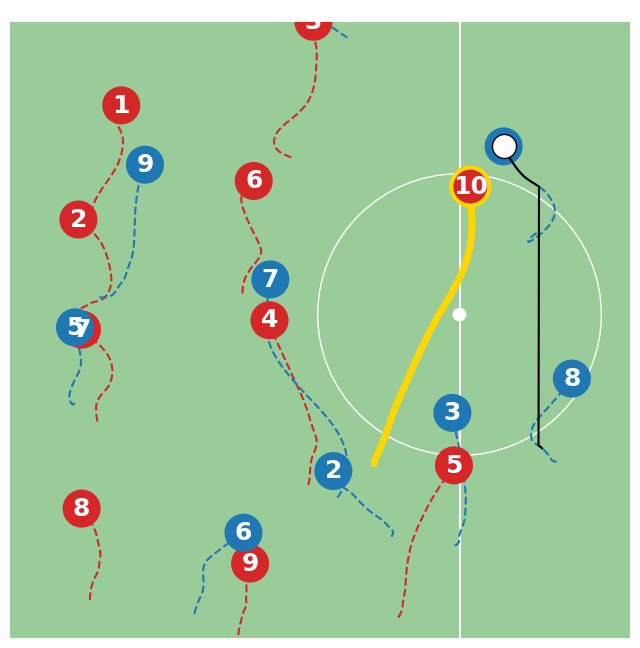}
        \caption{Pressing (PRS).}
        \label{fig:example_prs}
    \end{subfigure}
    \begin{subfigure}[t]{0.23\textwidth}
        \centering
        \includegraphics[width=\textwidth]{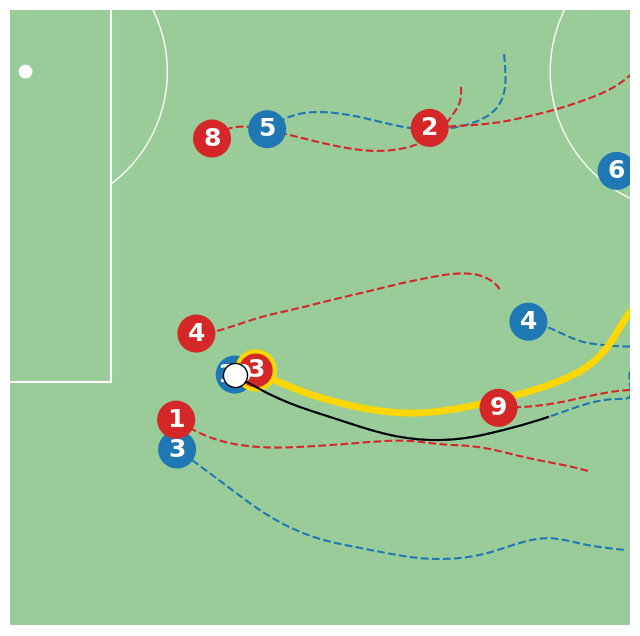}
        \caption{Chasing the Opponent (CTO).}
        \label{fig:example_cto}
    \end{subfigure}
    \caption{Instances of sprint categories. Note that the sprinter's team (colored in red) plays from left to right in every instance.}
\label{fig:examples}
\end{figure}

In this study, we adapt the taxonomy proposed by Ju et al. \shortcite{JuDHELB22} for labeling sprint categories to have more quantitative criteria. As advised by domain experts, we separate Overlapping (OVL) and Underlapping (UNL) and distinguish Move to Receive (MTR) from Exploiting Space (EXS). In addition, we created a new category named Chasing the Opponent with Ball (CTO), which originally blended in Pressing (PRS) and Covering (COV). See the resulting categories listed in Table~\ref{tab:rules} and their visualized examples in Figure~\ref{fig:examples}.

Based on this extended taxonomy, we construct a rule-based sprint-classifying engine to reduce the cost of annotation in Section \ref{se:human_annot} and assess the effectiveness of our deep learning classifier introduced in Section \ref{se:deep_model}. First, the engine regards a given sprint as \emph{attacking} if the sprinter's team has the ball for longer than either \SI{80}{\%} during the sprint interval or \SI{80}{\%} during the former half of the interval. On the contrary, it regards the sprint as \emph{defending} if the opposite team possesses the ball for longer than the same proportion of the time. The remaining sprints are unclassified and labeled as others (OTH).

\begin{figure}[htb]
    \centering
    \includegraphics[width=0.45\textwidth]{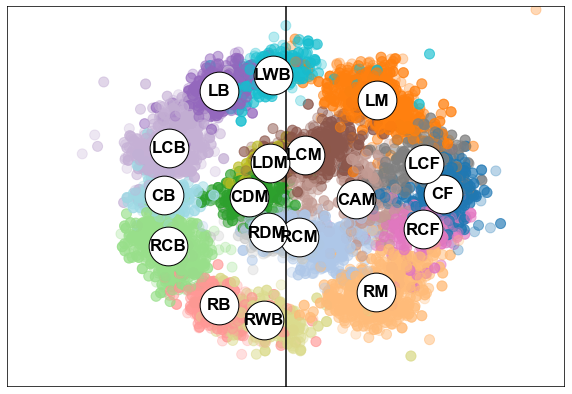}
    \caption{Distribution of 18 roles' mean locations per role period resulting from applying SoccerCPD to our dataset. The letters `L', `C', and `R' in the front of labels stand for `left', `central', and `right', respectively. Also, `(W)B', `(D/A)M', and `F' on the latter part signify `(wing-)back', `(defensive/attacking) midfielder', and `forward', respectively. For instance, `LCB' means `left center back'.}
    \label{fig:roles}
\end{figure}

\begin{table*}[h!]
  \caption{Description of sprint categories and their detection conditions used by the rule-based engine. Words written in italics are terms defined in Section~\ref{se:rule_engine}. Note that $(x,y)$ coordinates are aligned with the pitch so that the team attacks in the direction along which $x$ increases.}
  \label{tab:rules}
  \centering
  \renewcommand{\arraystretch}{1.1}
  \setlength{\tabcolsep}{3.5pt}
  \begin{tabularx}{\textwidth}{V{2.7cm}|V{6.3cm}|X}
    \toprule
    \multicolumn{1}{c|}{\textbf{Category}}       &
    \multicolumn{1}{c|}{\textbf{Description}}    &
    \multicolumn{1}{c}{\textbf{Conditions}}  \\
    \midrule
    \multicolumn{3}{c}{\textbf{In Possession (Attacking)}} \\
    \midrule
    Run with Ball (RWB)
        & The player runs with the ball.
        & The player has the ball for longer than \SI{40}{\%} of the period. \\
    Exploiting Space (EXS)
        & The player moves forward to receive the ball or to create/exploit a space (usually when the ball is behind).
        & (a) The player moves forward, and (b) the sprint either starts ahead or (c) ends more than \SI{3}{m} ahead of the ball. \\
    Penetration (PEN)
        & The player runs forward and overtakes/unbalances the opponents' defensive line.
        & When the sprint ends, (a) it heads for the \emph{scoring zone}, (b) the sprinter is in between the opponents' \emph{defensive line} and their end line, and (c) his/her \emph{goal side} is \emph{open}. \\
    Break into Box (BIB)
        & The player enters the opponents' penalty box to receive the ball from a cross.
        & (a) The sprint ends in the opponents' penalty box, and (b) a cross is expected (i.e., the teammate having the ball is at a flank) or actually occurs. \\
    Support Play (SUP)
        & The player behind the ball moves forward to support the teammate having the ball (usually getting close to the teammate).
        & (a) The player moves forward, and (c) the sprint starts behind the ball. \\
    Overlapping (OVL)
        & The player at the flank moves forward to run past a close teammate (usually having the ball) at a relatively central channel.
        & (a) The player moves forward, (b) the sprint starts behind the ball and (c) ends at a flank ahead of the ball, and (d) the player instantly takes a side role (i.e., LB, LWB, LM, RB, RWB, or RM) when the sprint ends. \\
    Underlapping (UNL)
        & The player moves from a flank behind to a forward half-space to run past a close teammate (usually having the ball) at the same flank.
        & (a) The player moves forward, (b) the sprint starts behind the ball and (c) ends at a flank ahead of the ball, and (d) the ball-possessing teammate at the end is located in between the sprint path and the near sideline. \\
    Move to Receive (MTR)
        & The player ahead of the ball moves sideward or backward to get close to the ball and receive a pass.
        & (a) The player moves backward, and (b) the distance between the player and the ball reduces during the sprint. \\
    \midrule
    \multicolumn{3}{c}{\textbf{Out of Possession (Defending)}} \\
    \midrule
    Closing Down / Pressing (PRS)
        & The player runs directly toward an opponent having/receiving the ball or a possible passing line of the opposite team to disturb their build-up.
        & (a) The \emph{target distance} decreases during the sprint and becomes smaller than \SI{5}{m} at the end or (b) the minimum distance between the sprint path and the opponents' \emph{potential passing lines} is smaller than \SI{3}{m}. \\
    Covering (COV)
        & The player runs backward to the team's defensive area to cover a space or close the goal side of an opponent.
        & (a) Both the player and his/her team \emph{return to defense}, and (b) the sprint ends in their \emph{defensive area}. \\
    Recovery Run (REC)
        & The player in the front moves back toward their own goal when the team is under attack on their side.
        & (a) Both the player and his/her team \emph{return to defense}, (b) the player's $x$-coordinate is always larger than that of the ball, and (c) the average difference between these $x$-coordinates is larger than \SI{10}{m}. \\
    Interception (INT)
        & The player tries to cut out a pass.
        & (a) The sprint meets an \emph{actual passing line} with an angle larger than $30^\circ$, and (b) the sprint interval is included in the extended pass interval with \SI{2}{s} margin. \\
    Chasing the Opponent with Ball (CTO)
        & The player chases an opponent running with the ball to steal it.
        & (a) The \emph{target} has the ball for longer than \SI{40}{\%} of the period, (b) his/her average speed is greater than \SI{15}{\kilo\meter\per\hour}, and (c) average \emph{target distance} is smaller than \SI{4}{m}. \\
    \midrule
    \multicolumn{3}{c}{\textbf{Common}} \\
    \midrule
    Push up Pitch (PUP)
        & The player at the defensive line moves forward to squeeze the gap between offense and defense or leave some opponents offside.
        & (a) The player belongs to the \emph{defensive line}, (b) the \emph{offside line} goes up for more than \SI{10}{m}, and (c) his/her average distance from the ball is larger than \SI{20}{m}. \\
    Others (OTH)
        & All remaining runs that are not categorized by the above.
        & The sprint does not fall under any of the above categories. \\
    \bottomrule
  \end{tabularx}
\end{table*}

Then, the engine further categorizes each sprint by the rules described in Table \ref{tab:rules}
based on the trajectories of players and the ball. For some categories (PEN, OVL, UNL, and PUP), it utilizes momentary roles of players detected by SoccerCPD \cite{KimKCYK22} that assigns 10 tactical roles among the 18 depicted in Figure~\ref{fig:roles} to 10 outfield players for every moment. Also, we adopt the following definitions when documenting the rules in Table~\ref{tab:rules}.

\begin{definition}[Goal side]
    The \emph{goal side} $G$ of a player $p$ is the polygon made by connecting $p$ and the opposite goal posts. $G$ is said to be \emph{open} if there is no opponent other than the goalkeeper in $G$ and said to be \emph{closed} otherwise.
\end{definition}

\begin{definition}[Target opponent]
    The \emph{target} of a defensive sprint is the opponent that is closest to the sprinter at the end of the sprint. Here we call the distance between the target and the player the \emph{target distance}.
\end{definition}

\begin{definition}[Passing line]
    An \emph{actual passing line} is defined as a line segment made by a real pass. Meanwhile, the \emph{potential passing lines} at a moment is the collection of line segments from a ball possessor $p$ to his/her neighbors. Here a teammate $q$ is said to be $p$'s \emph{neighbor} if Delaunay triangulation \cite{Delaunay34} applied to the locations of $p$'s teammates connects $p$ and $q$ by an edge.
\end{definition}

\begin{definition}[Offside line]
    The \emph{offside line} of a team at a moment is the line parallel to the end line and passing through the second rearmost player of the team including the goalie.
\end{definition}

\begin{definition}[Defensive line]
    The \emph{defensive line} of a team at a moment is a polyline made by connecting the players whose momentary roles are LB, LCB, CB, RCB, or RB and the feet of the perpendicular from each side defender (i.e., LB or RB) to the corresponding sideline.
\end{definition}

\begin{definition}[Defensive area]
    A \emph{defensive area} of a team at a moment is an area between their end line $L$ and the parallel line that is $d + \SI{20}{m}$ far from $L$ where $d$ is the average distance between $L$ and the players in the defensive line.
\end{definition}

\begin{definition}[Return to defense]
    A player/team is said to \emph{return to defense} if the player/team runs backward at an average vertical speed more than \SI{0.5}{\kilo\meter\per\hour}.
\end{definition}

For the case that a sprint satisfies the conditions of multiple categories, we prioritize intersecting categories as follows:
\begin{itemize}
    \item RWB $>$ BIB $>$ PEN $>$ EXS,
    \item RWB $>$ BIB $>$ UNL $>$ OVL $>$ SUP $>$ PUP,
    \item CTO $>$ INT $>$ PRS $>$ REC $>$ COV $>$ PUP.
\end{itemize}
The motivation for this prioritization is to maximize the recall of categories that are considered important by practitioners. For example, since penetrations are closely related to scoring opportunities, many coaches in professional clubs extract and analyze situations with penetrative passes or runs~\cite{Mychalczyk20,RahimianGGBT22}. Likewise, it is widely accepted that effective pressing was a tactical key of many successful clubs such as Jürgen Klopp's Liverpool FC and Gian Piero Gasperini's Atalanta BC. Accordingly, most contemporary analysts in high-level teams thoroughly investigate pressing situations that occurred during match-play \cite{AndrienkoABDFLW17,Robberechts19}. Thus, we prioritize PEN and PRS over other relatively common sprints such as EXS and COV, respectively, making sprints in those ``crucial'' categories detected with as few misses as possible.

\subsection{Constructing a Deep Learning Classifier}
\label{se:deep_model}
Although it could be deemed that the rule-based engine is sufficient to classify all sprints, there are many subtle cases where the engine fails to classify sprints into valid or correct categories. For instance, the rule-based engine depends on the distinguishment between attacking and defending situations, thereby often confusing categories of sprints that occurred during transition phases. Moreover, a single sprint may embody multiple intentions when there is an abrupt change of surrounding contexts but the detection algorithm introduced in Section~\ref{se:detection} cannot catch it to split the sprint interval. In this case, one solution is to assign a proper amount of probability to all the corresponding categories, but the rule-based engine inevitably puts the sprint into a single category.

To overcome this drawback, we propose a deep learning approach to the problem of classifying sprints. It uses the raw trajectories of all the players and the ball during each sprint instead of empirical conditions in Table~\ref{tab:rules} and returns a probability that the sprint belongs to each category. Thus, it does not suffer from the problem of the rule-based engine that is susceptible to handcrafted conditions and edge cases.

In regard to capturing the contextual information of a sprint, one technical consideration is that the trajectories of the sprinter and the other players should be processed in a permutation-invariant manner since the order of input players to the model is not important. Thus, inspired by a recent study that also handled trajectories in multi-agent sports \cite{KimCKYK23}, we employ Set Transformer \cite{LeeLKKCT19} to extract the context-aware embedding of a given situation while securing the permutation-invariance of input players.

To be specific, let $p_1$ be the considered sprinter and $P_1 = \{p_1, \ldots, p_n\}$ and $P_2 = \{p_{n+1}, \ldots, p_{2n}\}$ be his/her team and the opposing team, respectively. Let $\mathbf{x}^p_t$ be the feature values for each player $p$ at time $t$, including the 2D location, 2D velocity, speed, acceleration, and the 2D relative locations to the ball. Then, we encode the interaction between the sprinter and each team as the context-aware embeddings as follows:
\begin{IEEEeqnarray}{rCl}
    \mathbf{z}^1_t & = & \text{SetTransformer} (\mathbf{x}_t^{p_1}; \mathbf{x}_t^{p_2}, \ldots, \mathbf{x}_t^{p_n}), \\
    \mathbf{z}^2_t & = & \text{SetTransformer} (\mathbf{x}_t^{p_1}; \mathbf{x}_t^{p_{n+1}}, \ldots, \mathbf{x}_t^{p_{2n}}).
\end{IEEEeqnarray}

Then, we feed the sequence of the paired embeddings $\{(\mathbf{z}^1_t, \mathbf{z}^2_t)\}_{t=1}^T$ into a bidirectional GRU \cite{ChoVGBBSB14} to model its temporal dependencies:
\begin{equation}
\mathbf{h}_t = \text{Bi-GRU}(\mathbf{z}^1_t, \mathbf{z}^2_t; \mathbf{x}^{ball}_t)
\end{equation}
where $\mathbf{x}^{ball}_t$ is the ball location at $t$. Finally, a fully connected layer with the softmax activation converts the bidirectional GRU hidden state at the last time step into the probability vector $\hat{\mathbf{y}}_T$ that the sprint belongs to each of the $K$ categories:
\begin{equation}
\hat{\mathbf{y}} = (\hat{y}_1, \ldots, \hat{y}_K) = \text{softmax}(\text{FC}(\mathbf{h}_T)).
\end{equation}

Like many other classification models, the proposed classifier is trained by minimizing the cross-entropy loss between the probability vector $\hat{\mathbf{y}}$ and the ground truth $\mathbf{y}$ for each instance in the training dataset $\mathcal{D}$:
\begin{equation}
    \mathcal{L} = \frac{1}{|\mathcal{D}|} \sum_{\mathcal{D}} \sum_{k=1}^K y_k \log \hat{y}_k
\end{equation}
where the ground truth is obtained from the annotation process described in Section~\ref{se:human_annot}.

\section{Experiments}
This section elaborates on the processes of ground truth annotation and experiments conducted to train and evaluate the proposed deep learning classifier.

\subsection{Annotation of Ground Truth Labels}
\label{se:human_annot}
To train the deep learning classifier introduced in Section \ref{se:deep_model}, we integrated the predictions made by human annotators and the rule-based engine to obtain ground truth labels. In other words, we carried out human annotation based on the quantitative guideline described in Table \ref{tab:rules} and revised them with the assistance of the rule-based engine and domain experts.

First, we outsourced the annotation task to two labeling companies (referred to as A and B in this paper). They manually classified 5,105 sprints detected from 9 K League matches in total, where A and B independently annotated 4,005 and 3,046 with 1,946 sprints in common. The labels of 1,302 sprints among these 1,946 instances were consistent (i.e., inter-labeler reliability of \SI{66.9}{\%}), and 902 of them agreed with the classification results made by the rule-based engine.

Then, deeming these unanimous labels as ground truth, we revised the remaining 1,044 sprints with domain experts. Also, for the 3,159 sprints annotated by only one labeler, we corrected 1,307 of them in which the label and the rule-based classification result were not the same (A: 885 of 2,054, B: 422 of 1,100). Note that in the revision process, we carried out qualitative labeling using domain knowledge about the tactical meaning of each category rather than sticking to the quantitative rules. In summary, the domain experts revised 2,351 sprints among the total of 5,105, and the resulting labels were used as ground truth in training the deep learning model.

Table~\ref{tab:annot_acc} shows the classification accuracies of individual annotators and the engine. Since the annotators of both companies were not soccer experts, their accuracies were not as high as expected. Their work was still helpful since we could empirically fine-tune the conditions of the rule-based engine using the resulting labels. We leave experiments on the reliability of domain experts who revised the labels and the accuracy of the rule-based engine with varying conditions as future work.

\begin{table}[htb]
  \caption{Classification accuracies of the two labeling companies and the rule-based engine.}
  \label{tab:annot_acc}
    \begin{tabular}{l|ccc}
    \toprule
    Labeler     & Company A & Company B & Rule-based \\
    \midrule
    Common      & \makecell{1,228 / 1,946 \\ (\SI{63.10}{\%})}
                & \makecell{1,405 / 1,946 \\ (\SI{72.20}{\%})}
                & \makecell{1,523 / 1,946 \\ (\SI{78.26}{\%})} \\
    \midrule
    Total       & \makecell{2,603 / 4,005 \\ (\SI{64.99}{\%})}
                & \makecell{2,217 / 3,046 \\ (\SI{72.78}{\%})}
                & \makecell{4,027 / 5,105 \\ (\SI{78.88}{\%})} \\
    \bottomrule
    \end{tabular}
\end{table}

\begin{figure}[b!]
    \centering
    \includegraphics[width=0.48\textwidth]{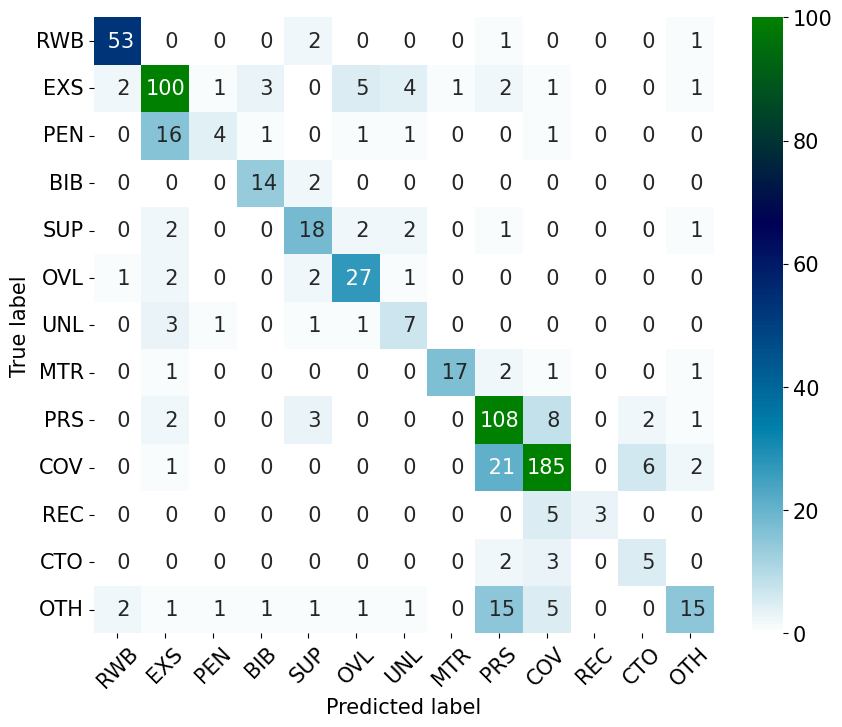}
    \caption{Confusion matrix for our deep learning classifier.}
    \label{fig:conf_mat}
\end{figure}

\begin{figure*}[t!]
    \centering
    \begin{subfigure}[t]{0.45\textwidth}
        \includegraphics[width=\textwidth]{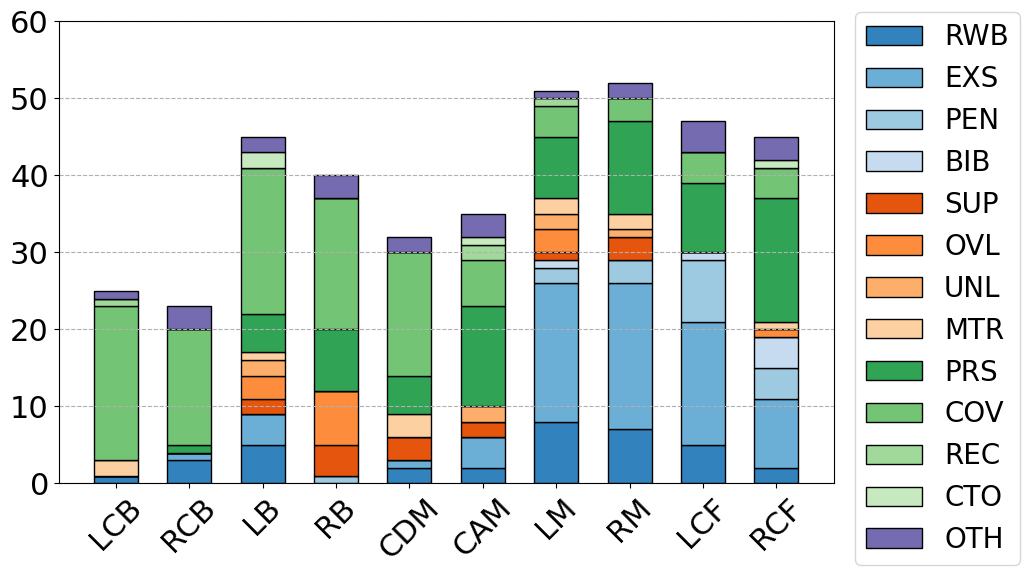}
        \caption{Team A with true categories.}
        \label{fig:counts_team1_true}
    \end{subfigure}
    \begin{subfigure}[t]{0.45\textwidth}
        \includegraphics[width=\textwidth]{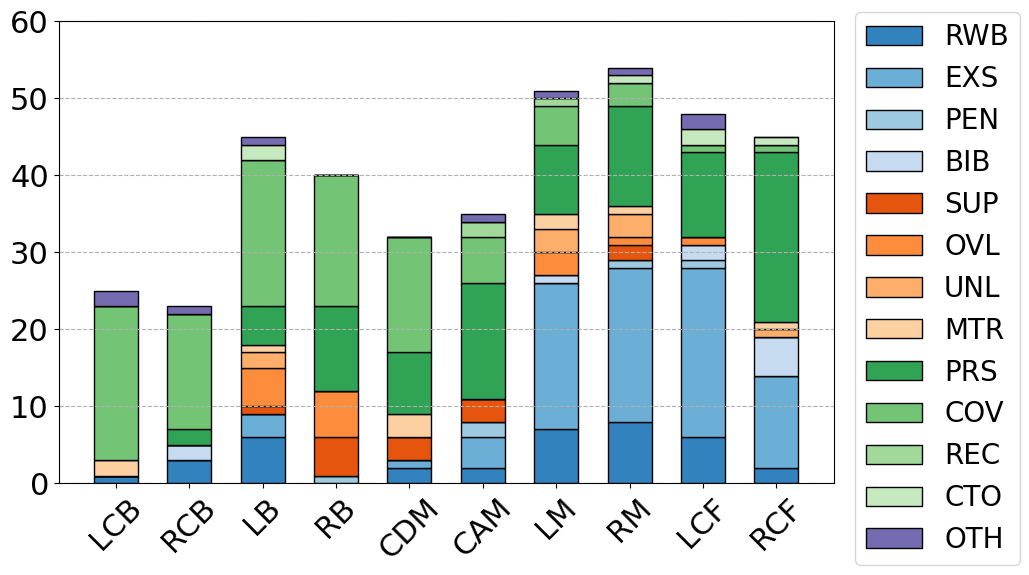}
        \caption{Team A with predicted categories.}
        \label{fig:counts_team1_pred}
    \end{subfigure}
    \begin{subfigure}[t]{0.45\textwidth}
        \centering
        \includegraphics[width=\textwidth]{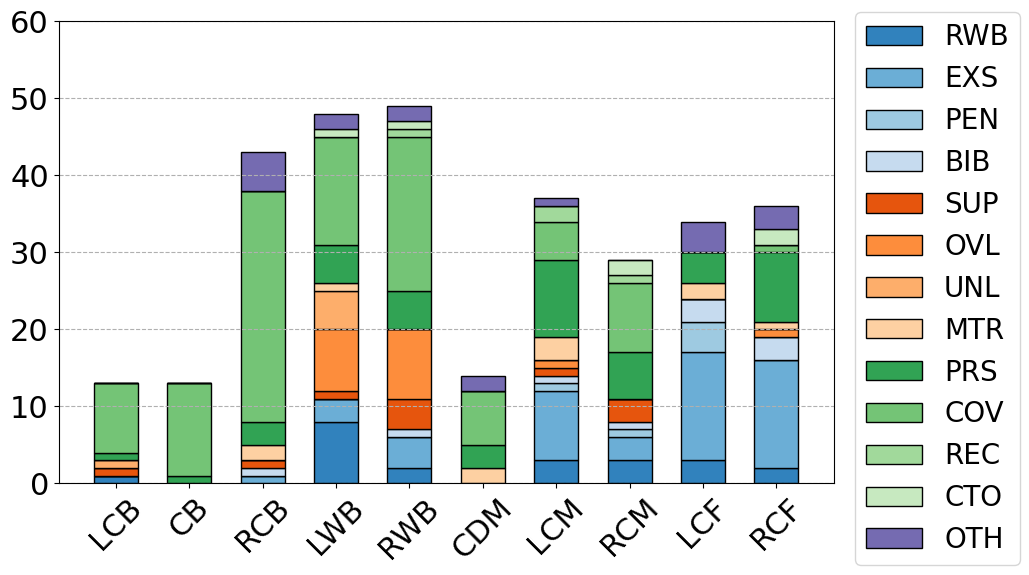}
        \caption{Team B with true categories.}
        \label{fig:counts_team2_true}
    \end{subfigure}
    \begin{subfigure}[t]{0.45\textwidth}
        \centering
        \includegraphics[width=\textwidth]{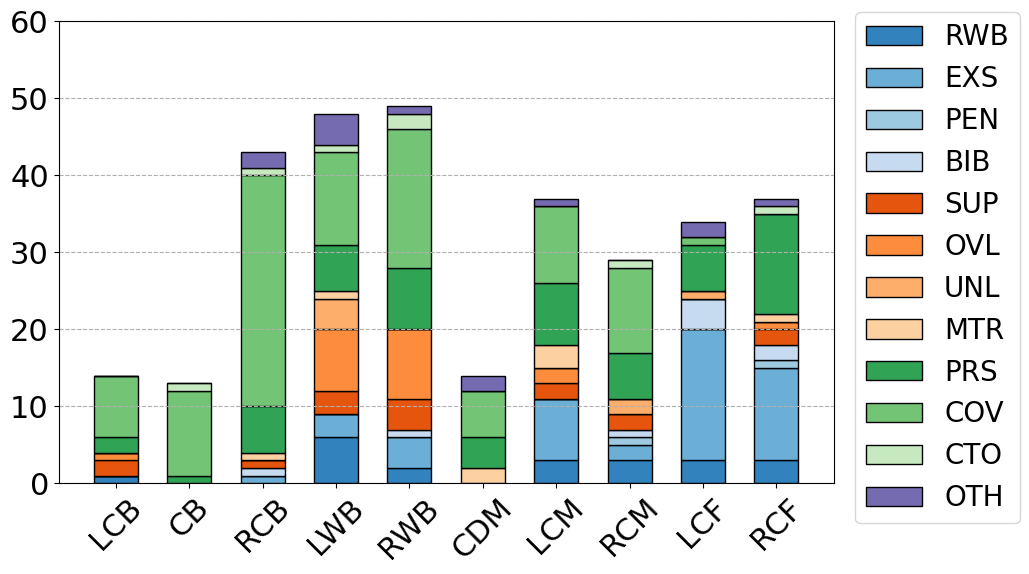}
        \caption{Team B with predicted categories.}
        \label{fig:counts_team2_pred}
    \end{subfigure}
    \caption{Role-by-role sprint counts of a match discretized by the true and predicted categories, respectively.}
\label{fig:sprint_counts}
\end{figure*}

\subsection{Performance of the Deep Learning Classifier}
\label{se:perf}
In this section, we conducted an experiment to evaluate the performance of our deep learning classifier. The dataset introduced in Section \ref{se:human_annot} consisting of 9 matches were split into 7-1-1 to allocate 3,877-512-716 sprints for training, validation, and test, respectively. That is, we trained the model using 3,877 sprints with ground truth labels resulting from Section \ref{se:human_annot} and compared the true and predicted categories of 716 sprints in the test dataset.

As a result, Figure~\ref{fig:conf_mat} shows the confusion matrix for the prediction of our model on the test dataset. The overall accuracy is \SI{77.65}{\%}, which is slightly lower than the rule-based engine with the accuracy of \SI{79.05}{\%}. However, most of the misclassifications come from two confusing pairs of categories, i.e., PEN-EXS and PRS-COV, implying room for improvement by upgrading the model architecture or adjusting the class imbalance. Also, note that there remain some irreducible biases in the ground truth labels since multi-intentional sprints have been also labeled as a single category. Though not systematically evaluated in this paper, we expect that the output probability values reflect their multiple intentions.

Future work will further investigate the output probabilities of the model with the annotation of multiple intentions for the corresponding sprints. We believe such investigation will substantiate the effectiveness of our deep learning-based approach with respect to the consideration in Section \ref{se:deep_model}.

\section{Use Cases}
In this section, we introduce some use cases of categorized sprints in practical scenarios.

\subsection{Aggregation of Categorized Sprints}
\label{se:agg}
The main motivation for contextualizing sprints in the field of sports science is to identify the physical demands of match-play decomposed by tactical purposes so that practitioners can design tailored training drills. Furthermore, one can figure out teams' and players' playing styles by analyzing their sprint tendencies. In this respect, not only the accuracy of individual sprint categories, it is also important that a classifier reliably estimates relevant metrics such as the number of occurrences or running distance grouped by the categories.

As an example, we counted sprints that occurred during the match in the test dataset for each category and player role. It should be noted that a player's tactical role may change over time even in a single match and the role significantly affects the player's sprint tendency. Hence, we applied SoccerCPD \cite{KimKCYK22} to the players' trajectories to detect their time-varying roles in the match and aggregated sprints by role instead of player identity. Unlike previous studies that either considered players with constant roles throughout the match \cite{JuDHELB22,JuDHELB23} or simply ignored the possibility of role change during the match \cite{CaldbeckD23}, this approach enables a precise analysis of general matches with time-varying roles.

\begin{figure*}[tb]
    \centering
    \begin{subfigure}[t]{0.29\textwidth}
        \includegraphics[width=\textwidth]{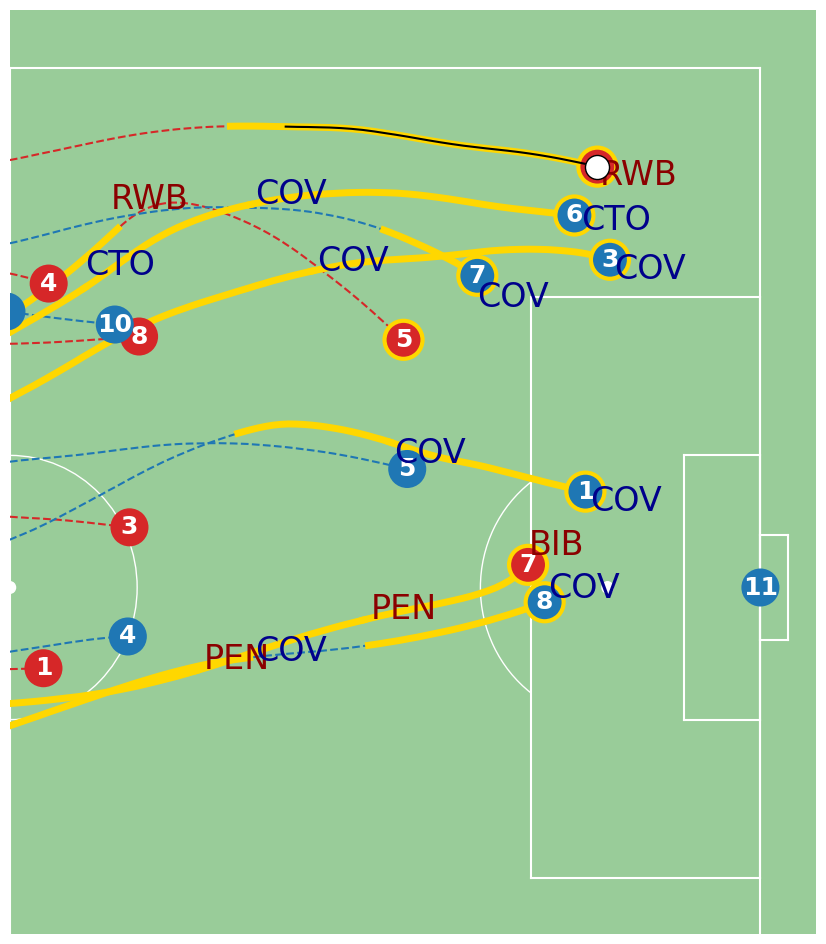}
        \caption{Query play.}
        \label{fig:query}
    \end{subfigure}
    \begin{subfigure}[t]{0.292\textwidth}
        \includegraphics[width=\textwidth]{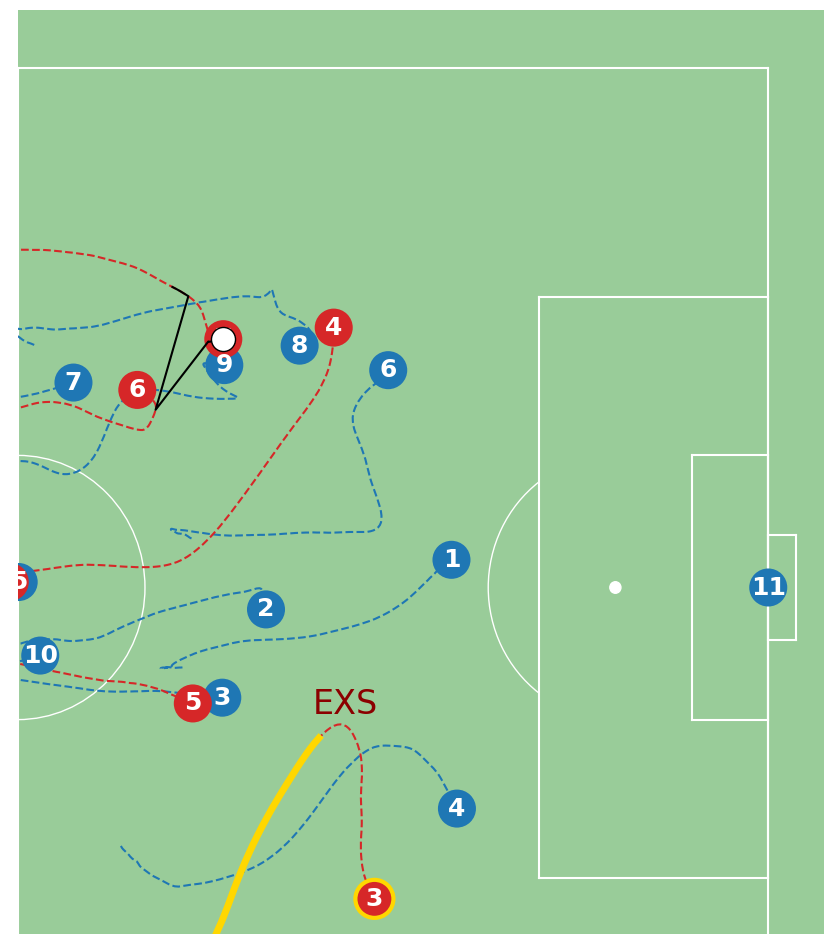}
        \caption{Vanilla play2vec.}
        \label{fig:play2vec}
    \end{subfigure}
    \begin{subfigure}[t]{0.29\textwidth}
        \centering
        \includegraphics[width=\textwidth]{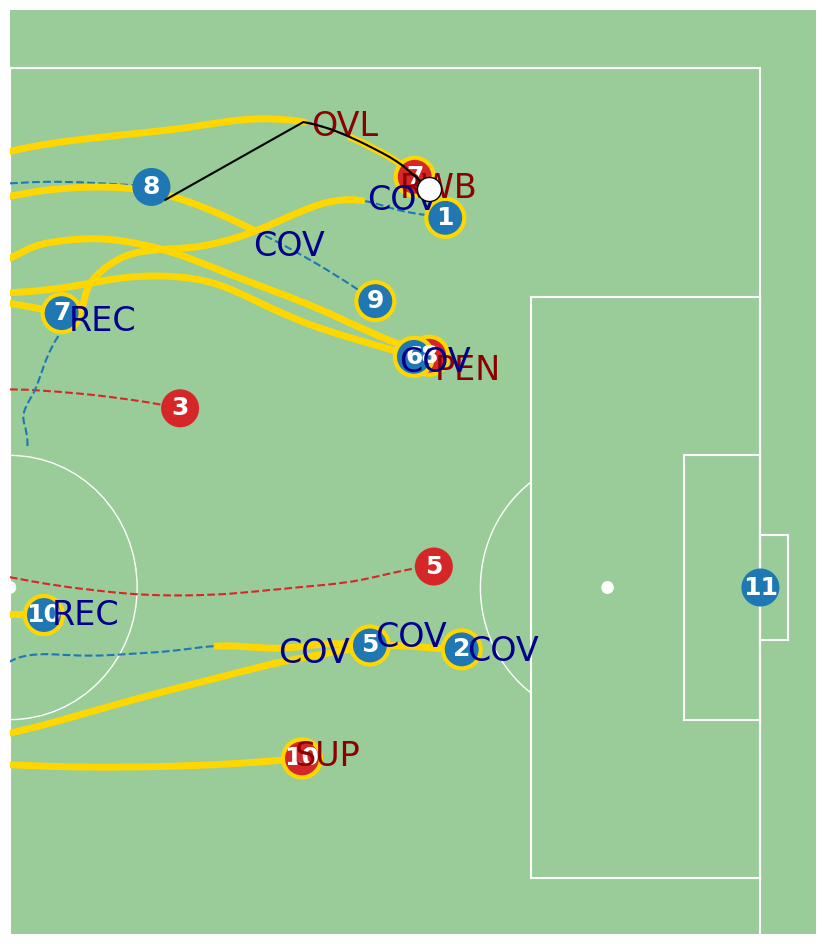}
        \caption{Keyword-guided play2vec.}
        \label{fig:keywords}
    \end{subfigure}
    \caption{An example play as a query and similar plays retrieved by play2vec with and without the keyword filtering.}
\label{fig:sim_play}
\end{figure*}

According to the resulting Figure~\ref{fig:sprint_counts}, the distributions made by predicted categories are similar to their true counterparts (i.e., \ref{fig:counts_team1_pred} to \ref{fig:counts_team1_true} and \ref{fig:counts_team2_pred} to \ref{fig:counts_team2_true}), implying that the prediction of the proposed deep learning classifier is precise enough to estimate the physical-tactical demands of matches. Besides, the prediction preserves important match semantics that the true distribution contains such as the following:
\begin{itemize}
    \item Overall, Team A performed more sprints than Team B. Especially, forward players (LM, RM, LCF, and RCF) in Team A made attacking sprints and PRS significantly more than those (LCM, RCM, LCF, and RCF) in Team~B.
    \item Team A adopted 4-4-2 with wingers (LM and RM) on both sides, while Team B used 3-5-2 without side midfielders. This leads to more sprints (OVL in particular) made by side defenders (LWB and RWB) in B than those in A.
    \item The RCB of Team B made way more COVs than other center backs (LCB and CB). A possible cause is that the team used an asymmetric formation by encouraging the RCB to participate in attacking.
\end{itemize}
These observations imply that though the proposed approach may not be \SI{100}{\%} accurate in classifying individual sprints, it can serve as a useful predictor of physical-tactical demands and contexts of soccer matches in practice.

\subsection{Keyword-Guided Similar Play Retrieval}
\label{se:sim_play}
Retrieving plays (i.e., fragments of a match) that are similar to a query play from a database is an important task in sports analytics for effective post-match analysis and coaching \cite{WangLCJ19}. Accordingly, several approaches have been proposed to tackle this task using trajectory data acquired in multi-agent sports. Sha et al. \shortcite{ShaLYCRM16,ShaLZKYS18} proposed methods to effectively calculate similarities between plays using tree-based alignment of trajectories. Wang et al. \shortcite{WangLCJ19} proposed play2vec that extracts a representation of playing sequences by combining Skip-Gram with Negative Sampling (SGNS) and a Denoising Sequential Encoder-Decoder architecture (DSED). Most recently, Wang et al. \shortcite{WangLC23} improved play2vec by introducing efficient search algorithms based on deep reinforcement learning and metric learning.

To help these frameworks retrieve more semantically relevant plays, we suggest another use case of the sprint categories as keywords to narrow down candidates. Namely, we first filter candidate plays with similar distributions of sprint categories to a given play and find close plays only among these candidates based on the similarity measure resulting from any play-retrieval framework. 

In order to demonstrate the effectiveness of this proactive filtering, we implemented play2vec \cite{WangLCJ19} and compared the outputs of the frameworks with and without the filtering. To this end, we first defined a \emph{play} as a series of players' movements from the beginning of a team's ball possession to the lose of possession or a pause of the game. Here we considered the team to lose the possession if the opposing team made three consecutive events afterwards. Then, we constructed a database consisting of 1,162 plays detected from the 9 matches and trained play2vec to extract their representations. Lastly, for a given query play, we either directly retrieved the closest play or filtered candidates with the same set of sprint categories before the similarity-based retrieval.

For a given query play shown in Figure \ref{fig:query} as an example, Figure \ref{fig:play2vec} is one that play2vec retrieves as the closest play without any keyword filtering, and Figure \ref{fig:keywords} is the closest among those containing PEN and RWB. In the domain-specific perspective, the latter is similar to the query in that a wide player is running with the ball and attackers in the central channel are penetrating or breaking into the threatening area. On the other hand, in the case of the former play, the overall spatial arrangement of players and the ball is similar to that of the query, but the context is different in terms of the intentions of running players. This is due to the limitation of the previous play-retrieval frameworks that similarities are calculated only based on the positional information in an unsupervised manner and implies the possibility of improving their quality of retrieval by putting more weight on semantically important attributes through the sprint category-based filtering. We leave systematic experiments for evaluating the effectiveness of leveraging sprint categories in similar play retrieval as future work.

\section{Conclusions}
In this study, we propose a quantified taxonomy and a deep learning classifier for automatically categorizing sprints according to their tactical purposes in soccer matches. In the future, we aim to improve the model with a richer dataset and conduct more systematic experiments on its applicability.

\section*{Acknowledgments}
This work was supported by the Ministry of Education of the Republic of Korea and the National Research Foundation of Korea (NRF-RS-2023-00208094). Also, the authors thank Jupyo Kim (Hwaseong FC, Republic of Korea) and Wonwoo Ju (Korea Football Association, Republic of Korea) for initiating a discussion on this study and revising sprint labels as domain experts.

\bibliographystyle{named}
\bibliography{ijcai24}

\end{document}